\documentclass{article} %
\usepackage{iclr2025_conference,times}

\usepackage{amsmath,amsfonts,bm}

\def\eqref#1{equation~\ref{#1}}

\def\1{\bm{1}}

\def\vs{{\bm{s}}}

\def\mH{{\bm{H}}}

\def\mK{{\bm{K}}}

\def\mM{{\bm{M}}}

\def\mQ{{\bm{Q}}}

\def\mV{{\bm{V}}}

\DeclareMathAlphabet{\mathsfit}{\encodingdefault}{\sfdefault}{m}{sl}
\SetMathAlphabet{\mathsfit}{bold}{\encodingdefault}{\sfdefault}{bx}{n}

\usepackage{hyperref}
\usepackage{url}

\newcommand{\ignore}[1]{}

\DeclareMathAlphabet{\mathbfit}{OML}{cmm}{b}{it}

\makeatletter
\DeclareRobustCommand\onedot{\futurelet\@let@token\@onedot}
\def\@onedot{\ifx\@let@token.\else.\null\fi\xspace}

\def\vs{\emph{vs}\onedot}

\def\method{DuoAttention\xspace}

\newcommand{\myparagraph}[1]{\vspace{0pt}\paragraph{#1}}

\usepackage{hyperref}

\usepackage{times}
\usepackage{latexsym}
\usepackage{amsmath}

\usepackage[T1]{fontenc}
\usepackage[utf8]{inputenc}

\usepackage{microtype}
\usepackage{bbding}
\usepackage{color,xcolor}
\usepackage{epsfig}
\usepackage{graphicx}

\usepackage{adjustbox}
\usepackage{array}
\usepackage{booktabs}
\usepackage{colortbl}
\usepackage{wrapfig}
\usepackage{hhline}
\usepackage{multirow}
\usepackage{subcaption} %

\usepackage{amsmath,amsfonts,amssymb}
\usepackage{bm}
\usepackage{nicefrac}
\usepackage{microtype}
\usepackage{mathtools}

\usepackage{changepage}
\usepackage{extramarks}
\usepackage{fancyhdr}
\usepackage{lastpage}
\usepackage{setspace}
\usepackage{soul}
\usepackage{xspace}

\usepackage{url}

\usepackage{enumerate}
\usepackage{enumitem}  %
\usepackage{titlesec}

\usepackage{makecell}

\usepackage{pifont} %

\usepackage{amsthm}
\usepackage{float}

\title{\method: Efficient Long-Context LLM Inference with Retrieval and Streaming Heads}

\author{\textbf{Guangxuan Xiao}$^{1}$
\thanks{Part of the work done during an internship at NVIDIA.}
\quad
\textbf{Jiaming Tang}$^{1}$
\quad
\textbf{Jingwei Zuo}$^{2}$
\quad
\textbf{Junxian Guo}$^{1,3}$
\\
\textbf{Shang Yang}$^{1}$
\quad
\textbf{Haotian Tang}$^{1}$
\quad
\textbf{Yao Fu$^{4}$}
\quad
\textbf{Song Han$^{1,5}$} \\
$^1$ MIT
\quad
$^2$ Tsinghua University
\quad
$^3$ SJTU
\quad
$^4$University of Edinburgh
\quad
$^5$ NVIDIA
\\ \texttt{\url{https://github.com/mit-han-lab/duo-attention}}
}

\iclrfinalcopy %
\begin{document}

\maketitle

\begin{abstract}
Deploying long-context large language models (LLMs) is essential but poses significant computational and memory challenges.
Caching all Key and Value (KV) states across all attention heads consumes substantial memory.
Existing KV cache pruning methods either damage the long-context capabilities of LLMs or offer only limited efficiency improvements.
In this paper, we identify that only a fraction of attention heads, a.k.a, \emph{Retrieval Heads}, are critical for processing long contexts and require full attention across all tokens.
In contrast, all other heads, which primarily focus on recent tokens and attention sinks--referred to as \emph{Streaming Heads}--do not require full attention.
Based on this insight, we introduce \method, a framework that only applies a full KV cache to retrieval heads while using a light-weight, constant-length KV cache for streaming heads, which reduces both LLM's decoding and pre-filling memory and latency without compromising its long-context abilities.
\method uses a lightweight, optimization-based algorithm with synthetic data to identify retrieval heads accurately.
Our method significantly reduces long-context inference memory by up to 2.55$\times$ for MHA and 1.67$\times$ for GQA models while speeding up decoding by up to 2.18$\times$ and 1.50$\times$ and accelerating pre-filling by up to 1.73$\times$ and 1.63$\times$ for MHA and GQA models, respectively, with minimal accuracy loss compared to full attention.
Notably, combined with quantization, \method enables Llama-3-8B decoding with 3.3 million context length on a single A100 GPU.
Code is provided in the \href{https://github.com/mit-han-lab/duo-attention}{link}.
\end{abstract}

\section{Introduction}
Large language models (LLMs)~\citep{touvron2023llama,touvron2023llama2,openai2023gpt4,black2022gptneox20b} are at the forefront of the AI revolution, powering advanced applications such as multi-round dialogues~\citep{schulman2022chatgpt,alpaca,vicuna2023}, long document summarization~\citep{goyal-durrett-2020-evaluating,zhang2023benchmarking}, and tasks involving mixed modalities like visual and video understanding~\citep{liu2023visual,lin2023videollava}.
These applications often require processing extensive numbers of contextual tokens; for instance, summarizing the entire Harry Potter series could involve analyzing approximately one million tokens. The challenge intensifies with visual language models (VLMs), where a single 224$\times$224 image corresponds to 256 tokens~\citep{liu2023visual}, and a three-minute video at 24 FPS generates around 1.1 million tokens.

A critical issue in deploying LLMs in such applications is the long-context inference problem. The full attention mechanism demands that all tokens attend to every previous token for accurate representation, resulting in linearly increasing decoding and quadratically increasing pre-filling latency as the sequence length grows. Additionally, the Key-Value (KV) Cache technique, which stores keys and values from all preceding tokens, causes memory usage to scale linearly with context length. As sequences lengthen, memory is increasingly consumed by the KV cache, placing a significant computational burden on the attention mechanism. For instance, in the Llama-3-8B~\citep{dubey2024llama3herdmodels} model architecture, serving with FP16 KV cache for 1 million tokens would require at least 137 GB of memory—exceeding the capacity of a single 80GB GPU. Additionally, the latencies of pre-filling and decoding with such large contexts are significant, posing substantial challenges to the effective use of LLMs in long-context scenarios.

Despite numerous efforts to overcome the challenges of attention mechanisms in long-context inference, significant computational and memory issues persist. Architectural modifications, such as Grouped-Query Attention (GQA)\citep{ainslie2023gqa}, require model pre-training and fail to reduce computational costs. Linear Attention methods~\citep{gu2023mamba,poli2023hyenahierarchylargerconvolutional}, while less demanding in terms of computation and memory, often underperform in long-context scenarios compared to Transformer models. 
Approximative attention methods, such as H$_2$O~\citep{zhang2023h2o}, StreamingLLM~\citep{xiao2023streamingllm}, TOVA~\citep{oren2024transformers}, and FastGen~\citep{ge2024model}, often compromise accuracy in long-context applications and are incompatible with essential KV cache optimization techniques like GQA.
KV cache quantization~\citep{liu2024kivi,hooper2024kvquant}, although useful, does not reduce the computation time of the attention mechanism. System-level optimizations, including FlashAttention~\citep{dao2022flashattention,dao2023flashattention2}, FlashDecoding~\citep{hong2024flashdecoding}, and PagedAttention~\citep{kwon2023efficient}, while effective, do not reduce the KV cache size and still require significant computation for extended contexts. These limitations emphasize the need for further advancements to deploy models that handle million-level context lengths.

\begin{figure}[t]
    \centering
    \includegraphics[width=\textwidth]{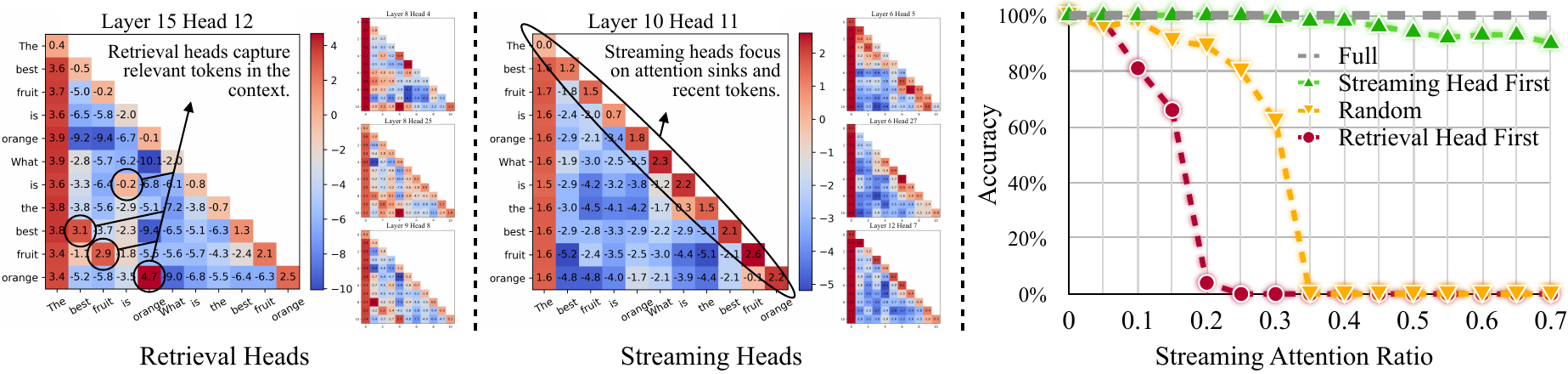}
    \caption{\small Visualization of attention maps in the Llama-2-7B model for the sentence "\textit{The best fruit is orange. What is the best fruit? Orange.}" shows the distinct roles of \emph{retrieval heads} (e.g., Layer 15, Head 12) and \emph{streaming heads} (e.g., Layer 10, Head 1). On the left, retrieval heads capture contextually relevant tokens such as "best," "fruit," and "orange," which are crucial for processing long-context information and, therefore, require a full KV cache. In the middle, streaming heads primarily focus on initial and recent tokens without emphasizing past contextual relevance. On the right, the impact of limiting attention to the sink and recent tokens on long-context passkey retrieval accuracy is shown: modifying retrieval heads severely damages performance, while constraining streaming heads has minimal impacts.
    }
    \label{fig:attn_vis}
\end{figure}

In this paper, we introduce a key observation that attention heads in LLMs can be categorized into two distinct types: Retrieval Heads~\citep{wu2024retrieval} and Streaming Heads, as shown in Figure~\ref{fig:attn_vis}. \emph{Retrieval Heads}, which represent only a fraction of the total, are crucial for processing long contexts and require full attention across all tokens. In contrast, the majority of attention heads, termed \emph{Streaming Heads}, primarily focus on recent tokens and attention sinks~\citep{xiao2023streamingllm}, and can operate effectively with a reduced KV cache that includes only recent tokens and attention sinks.

Building on the dichotomy of retrieval and streaming heads, we propose \method, a general, straightforward, and easily integrated approach that significantly accelerates both LLM's decoding and pre-filling and reduces memory footprints, particularly in long-context scenarios.
The core innovation of \method is a lightweight, optimization-based procedure that identifies non-compressible retrieval heads using synthetic datasets. Unlike existing methods that rely on attention pattern profiling~\citep{wu2024retrieval,ge2024model,tang2024razorattentionefficientkvcache}, \method directly measures output deviation resulting from token dropping, achieving higher compression rates and improved deployment efficiency.
\method is designed with simplicity and efficiency in mind: each Transformer layer has two KV caches— a full KV cache for crucial retrieval heads and a constant KV cache for streaming heads, which stores only attention sinks and recent tokens. This design allows \method to dramatically reduce memory usage and improve decoding speed in models like Llama-2/3 and Mistral, achieving up to 2.55$\times$ for MHA and 1.67$\times$ for GQA models while speeding up decoding by up to 2.18$\times$ and 1.50$\times$ and accelerating pre-filling by up to 1.73$\times$ and 1.63$\times$ for MHA and GQA models, respectively, with minimal accuracy loss compared to full attention.

Moreover, \method is fully compatible with important optimization techniques like GQA and quantization. We show that when combined with 8-bit weight 4-bit KV cache quantization, \method enables a Llama-3-8B model to handle up to 3.3 million contextual tokens measured on a single A100 GPU, achieving a 6.4$\times$ capacity increase compared to standard full attention FP16 deployments. \method paves the way for deploying LLMs in applications requiring million-level context handling.

\section{\method}
\label{sec:method}
\begin{figure}[t]
    \centering
    \includegraphics[width=1.0\linewidth]{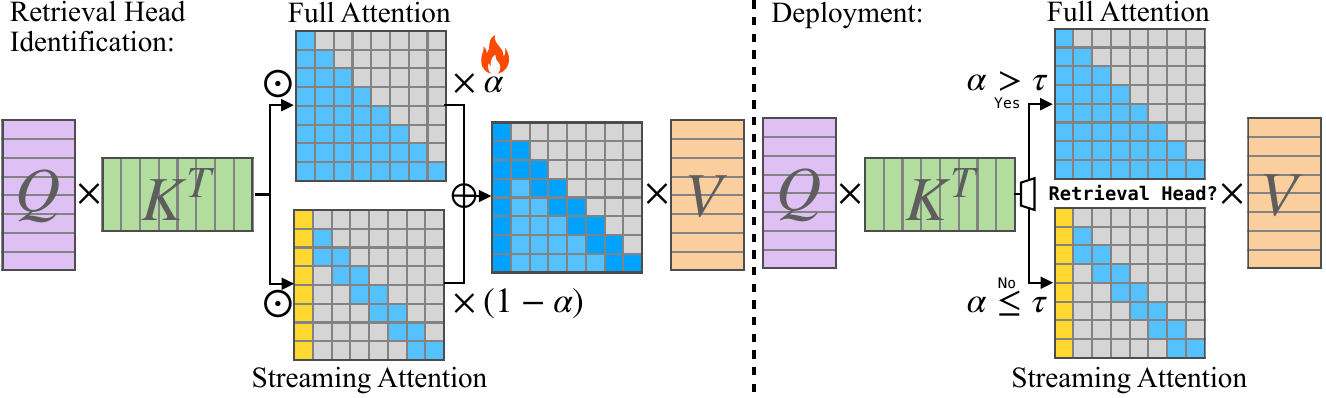}
    \caption{\small \textbf{Overview of \method:} (1) In the retrieval head identification phase, we assign a trainable gate value, $\alpha$, to each attention head, which blends the outputs of full attention and streaming attention. The training objective is to optimize these values to minimize the deviation from the full attention model's output, while simultaneously applying a regularization loss to encourage lower gate values. This training phase is efficient, requiring only the gate values to be trainable—leaving all other model parameters frozen—thus allowing it to be completed within several hours on an 8  GPU node. (2) During deployment, these gate values are binarized to classify heads as either retrieval or streaming based on a threshold $\tau$. Retrieval heads, identified by a gate value above the threshold, use full attention, caching the KV pairs for all tokens. In contrast, streaming heads cache only the KV pairs of recent tokens and attention sinks.
    }
    \label{fig:method}
\end{figure}
\subsection{Retrieval and Streaming Heads}
\label{sec:retrieval_vs_streaming}
\myparagraph{Retrieval Heads} In Transformer-based LLMs, attention heads exhibit distinct and consistent patterns, reflecting their specialized functionalities~\citep{clark-etal-2019-bert,xiao2023streamingllm,wu2024retrieval}. Figure~\ref{fig:attn_vis} visualizes two types of attention heads in the Llama-2-7B-32K-Instruct model using the sentence "\textit{The best fruit is orange. What is the best fruit? Orange}". The left panel highlights an attention head that emphasizes relevant tokens during decoding; for instance, the first occurrence of "best fruit" is accentuated while decoding the second "best fruit," and the initial "orange" is highlighted when inferring the second "orange." These attention heads, which we term \emph{Retrieval Heads}, are crucial for context processing as they capture contextually relevant tokens. Compressing the KV cache for retrieval heads would lead to the loss of vital contextual information, and thus they require full attention across all tokens.

\myparagraph{Streaming Heads}In contrast, the attention head depicted in the middle panel of Figure~\ref{fig:attn_vis} primarily attends to recent tokens and attention sinks~\citep{xiao2023streamingllm}, without highlighting earlier relevant tokens in the context. We refer to these as \emph{Streaming Heads}. Compressing the KV cache for Streaming Heads is feasible because dropping the unattended middle tokens does not significantly alter the attention output. Therefore, streaming heads can be optimized by retaining only the KV states of attention sinks and recent tokens, without compromising the model's ability to manage long contexts.

\myparagraph{Impact of Token Pruning on Retrieval and Streaming Heads} The right panel of Figure~\ref{fig:attn_vis} shows a preliminary passkey retrieval experiment, showing that the model's performance drops significantly when the middle tokens in the KV cache of retrieval heads are pruned, i.e., replaced with streaming attention. In contrast, removing the middle tokens for streaming heads has no significant impact on passkey retrieval accuracy. This observation indicates that we can enhance computational efficiency without sacrificing the model’s long-context capabilities: By dropping middle tokens for streaming heads while keeping full attention for retrieval heads, we reduce the memory demands of streaming heads to $O(1)$, thereby improving the efficiency of processing long contexts.

\begin{figure*}[t]
    \centering
    \begin{minipage}{0.41\textwidth}
        \centering
        \includegraphics[width=0.7\textwidth]{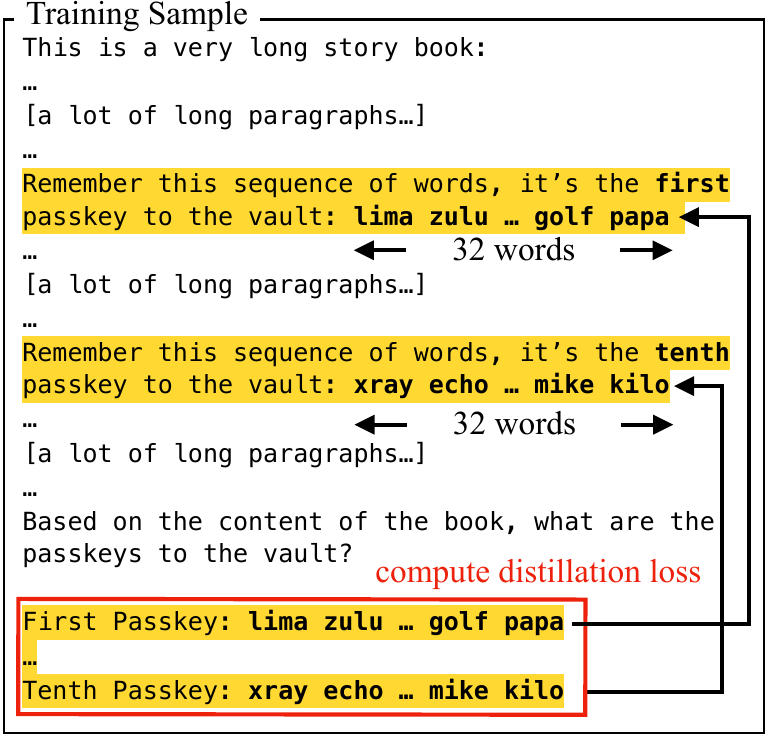}
        \captionof{figure}{\small Example from the synthetic dataset used to identify retrieval heads. We embed ten 32-word passkeys within a long text and ask the model to recall these passkeys. Distillation loss is calculated solely on the passkeys.}
        \label{fig:synthetic_data}
    \end{minipage}\hfill
    \begin{minipage}{0.57\textwidth}
        \centering
        \includegraphics[width=\linewidth]{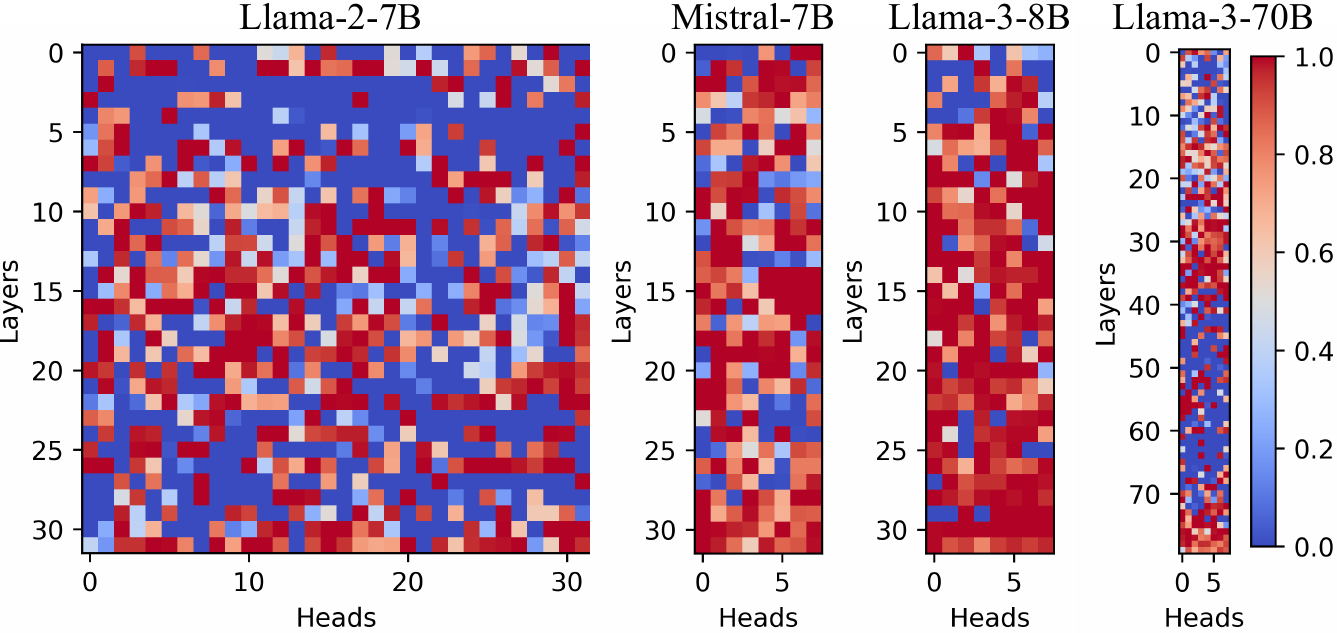}
        \vspace{-0.8em}
        \captionof{figure}{\small Optimized gate values of four LLMs. Llama-2-7B uses MHA with 32 heads per layer, while Mistral and Llama-3 models use GQA with 8 heads per layer. Retrieval heads have higher scores. MHA models have a lower ratio of retrieval heads compared to GQA models.}
        \label{fig:trained_mask}
    \end{minipage}
\end{figure*}
\subsection{Optimization-Based Identification of Retrieval Heads}
\label{sec:identifying}
\myparagraph{Definition of Retrieval Heads}Section~\ref{sec:retrieval_vs_streaming} qualitatively defines retrieval and streaming heads, but for precise identification, we need a concrete and quantitative definition. In this paper, we define ``retrieval heads'' as the attention heads that:
\begin{center}
    \emph{significantly alter model outputs when restricted to recent tokens and attention sinks.}
\end{center}
We use this criterion to distinguish retrieval heads from streaming heads. Note that this definition differs from existing works~\citep{ge2024model,wu2024retrieval,tang2024razorattentionefficientkvcache} that rely solely on attention scores to identify retrieval heads, which overlook 1) the end-to-end impact of compressing the KV cache for specific attention heads, 2) the role of value states, and 3) the variability of attention distributions across layers and heads. In contrast, our definition directly measures output deviation, allowing us to identify attention heads crucial for long-context processing, \emph{even when they are not apparent in attention scores}. We support this argument with ablation studies presented in Section~\ref{sec:ablation}.

\myparagraph{Optimization-based Identification} We employ an optimization-based approach to identify retrieval heads, drawing inspiration from prior work in CNN filter pruning~\citep{Liu2017learning}, as illustrated in Figure~\ref{fig:method}. First, we assign a gate value $\alpha_{i,j}$, to each key-value (KV) head in the LLM. This value intuitively represents the importance of the $j$-th KV head in layer $i$ for processing long-context information. Note that in models using GQA, one KV head can be associated with multiple attention heads, and our method accounts for the KV cache compression of an entire group of attention heads.

Our optimization-based identification method directly assesses the impact of compressing the KV cache with only sink and recent tokens for each KV head. We begin by initializing the gate value $\alpha_{i,j}\in[0,1]$ for each head at 1, assuming that all heads initially serve as retrieval heads. These gate values are then optimized, with the LLM’s parameters remaining fixed, limiting the number of trainable parameters to $N\times H$ and preventing the impact to the model's original abilities.

During the forward pass, we combine the outputs of full and streaming attention (which attends only to sink and recent tokens) for each KV head, using the gate value as the mixing weight:
\begin{equation*}
    \small
    \texttt{attn}_{i,j} = \alpha_{i,j} \cdot \texttt{full\_attn} + (1 - \alpha_{i,j}) \cdot \texttt{streaming\_attn}
\end{equation*}
where the attention calculations are defined as:
\begin{align*}
    \small
    \vspace{-0.5em}
    \texttt{full\_attn} &= \text{softmax}(\mQ\mK^T \odot \mM_{\text{causal}})\mV , \\
    \texttt{streaming\_attn} &= \text{softmax}(\mQ\mK^T \odot \mM_{\text{streaming}})\mV,
    \vspace{-0.5em}
\end{align*}
where $\mM_{\text{causal}}$ is the causal attention mask (a lower triangular matrix), and $\mM_{\text{streaming}}$ represents a $\Lambda$-like mask~\citep{han2023lminfinite,xiao2023streamingllm} that attends only to recent and initial tokens.

\myparagraph{Synthetic Dataset for Identifying Retrieval Heads}However, relying solely on natural language modeling objectives is insufficient for identifying retrieval heads because the supervision signal in natural text that requires inference over long spans is sparse, and most tokens can be inferred using local context. To address this, we design a synthetic dataset specifically aimed at enhancing the model's long-context retrieval capabilities, allowing us to effectively identify which KV heads can be compressed without compromising the model's performance.
As depicted in Figure~\ref{fig:synthetic_data}, we create a passkey-retrieval dataset by embedding ten randomly generated passkey sequences of $s$ tokens in ten random locations within a very long context ($s=32$ in experiments). The model is then tasked with recalling these ten sequences at the end of the context.

\myparagraph{Training and Loss Functions} We optimize the distillation loss, which is the L2 difference between the e last hidden state of the full attention model ($\mH_{\text{full}}$) and those of the model using \method ($\mH_{\text{mixed}}$), focusing only on the last $l$ passkey tokens in the entire inputs with $T$ tokens:
\begin{equation}
    \mathcal{L}_{\text{distill}} = \frac{1}{N} \sum_{i=1}^{N} \sum_{j=T-l+1}^{T} ( \mH_{\text{full}}^{(i)}[j] - \mH_{\text{mixed}}^{(i)}[j] )^2
\end{equation}
Our synthetic dataset ensures that every supervision signal is relevant to the final compression strategy, making the process lossless in terms of information retrieval accuracy. It proves to be more effective than using natural language modeling alone (see ablation studies in Section~\ref{fig:ablations}).
We use the L1 regularization term (a.k.a, Lasso~\citep{tibshirani96regression}) to encourage sparsity in the gate values:
\begin{equation}
    \mathcal{L}_{\text{reg}} = \sum_{i=1}^{L} \sum_{j=1}^{H} \left| \alpha_{i,j} \right|.
\end{equation}
The final training loss is a combination of the distillation loss and the regularization loss, weighted by a hyperparameter $\lambda$, which we set as 0.05 in our experiments:
\begin{equation}
    \mathcal{L} = \mathcal{L}_{\text{distill}} + \lambda \mathcal{L}_{\text{reg}}.
\end{equation}
Since the total number of trainable parameters is only thousands of floating-point numbers, this optimization process is fairly fast, with only 2,000 steps needed. All training experiments in our paper can be conducted on 8$\times$NVIDIA A100 GPU servers.

\begin{figure}[t]
    \centering
    \includegraphics[width=1.0\linewidth]{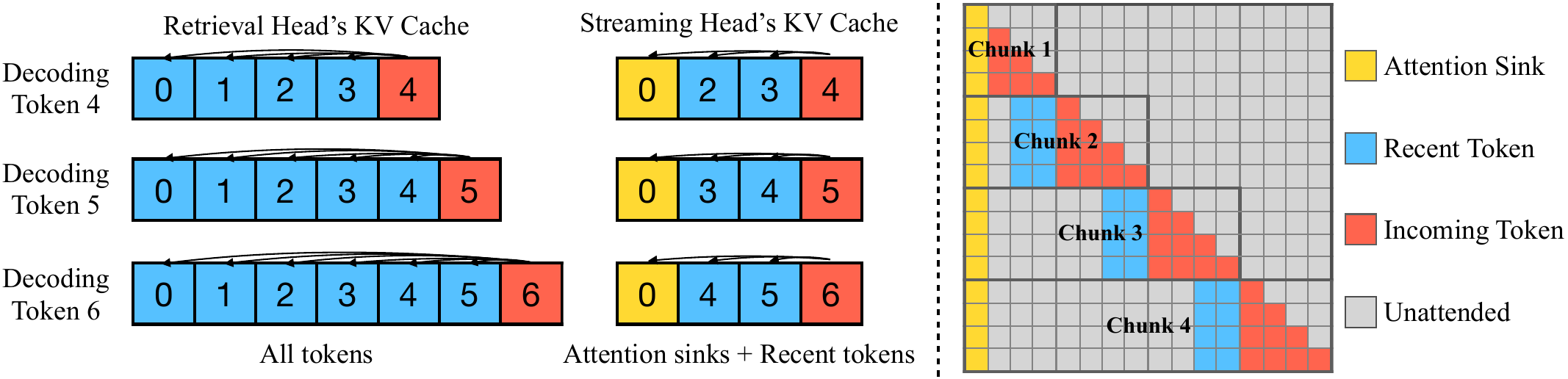}
    \caption{\small \textbf{Decoding (left) and Chunked Pre-filling (right) Processes in \method:} (1) The retrieval heads' KV cache stores all tokens, while the streaming heads' KV cache retains only recent tokens and attention sinks, ensuring constant memory usage. (2) The chunked pre-filling process of \method's streaming heads on a 16-token sequence, with one attention sink, two recent tokens, and a chunk size of 4. \method's streaming heads have linear time and constant memory complexity during long sequence pre-filling.
    }
    \label{fig:decoding_prefilling}
\end{figure}
\subsection{Deploying LLMs with \method}
\myparagraph{Binarizing Attention Implementations} At inference time, we apply full attention exclusively to the designated retrieval heads, identified using the optimized gate values from the training phase (as shown in Figure~\ref{fig:trained_mask}). We binarize the attention policy for each head based on a threshold $\tau$, determined by a specified sparsity quantile, to differentiate between retrieval heads and streaming heads:
\begin{equation}
\small
\vspace{-0.5em}
\texttt{attn}_{i,j} = \begin{cases}
\texttt{full\_attn} & \text{if } \alpha_{i,j} > \tau \\
\texttt{streaming\_attn} & \text{otherwise}
\end{cases}
\vspace{-0.5em}
\end{equation}

\myparagraph{Reordering Attention Heads} Before deployment, we preprocess the model by reordering the output channels of the Query, Key, and Value projection weights according to the attention head assignments. This reordering groups retrieval heads and streaming heads into two distinct, consecutive clusters, allowing for efficient slicing and concatenation operations when managing the KV cache for these two types of heads within a layer, rather than relying on scattering and gathering operations.

\myparagraph{Decoding} As shown in Figure~\ref{fig:decoding_prefilling}, we allocate two KV caches for each layer in the LLM during decoding: one for retrieval heads, which stores all past Keys and Values, and another for streaming heads, which stores only attention sinks and recent tokens, maintaining a constant size. When a new token is processed, its query, key, and value vectors are split along the head dimension to compute full attention for retrieval heads and streaming attention for streaming heads. The results are then concatenated along the head dimension for the output projection.

\myparagraph{Chunked Pre-filling}
\label{sec:prefilling}
We use FlashAttention-2~\citep{dao2023flashattention2} to pre-fill the KV caches for both retrieval and streaming heads. In long-context LLMs, chunked pre-filling is a common practice~\citep{agrawal2023sarathiefficientllminference,kwon2023efficient}, dividing the prompt into fixed-length chunks to pre-fill the KV cache. This technique significantly reduces peak memory usage (see Table~\ref{fig:efficiency_prefilling}) by lowering the peak intermediate activation size in linear layers from sequence length to chunk size.
\method is fully compatible with chunked pre-filling, and the streaming heads' pre-filling in \method can be achieved with linear time and constant memory complexity, \emph{without} requiring specialized kernels. As shown in Figure~\ref{fig:decoding_prefilling}, once a layer's KVs are computed, the streaming head's KV cache is immediately pruned to keep only the sink and recent tokens. The next chunk of incoming tokens will only attend to a constant number of contextual tokens during pre-filling. Let $L$ represent the sequence length and $K$ the chunk size. The pre-filling time complexity for streaming heads is optimized from $O(L^2)$ to $O(LK)$, and the memory complexity is reduced from $O(L)$ to $O(K)$.

It’s important to note that \method's design is well-suited for batch operations, which can further enhance LLM efficiency in serving scenarios with large batch sizes.
\begin{figure}[t]
    \centering
    \includegraphics[width=1.0\linewidth]{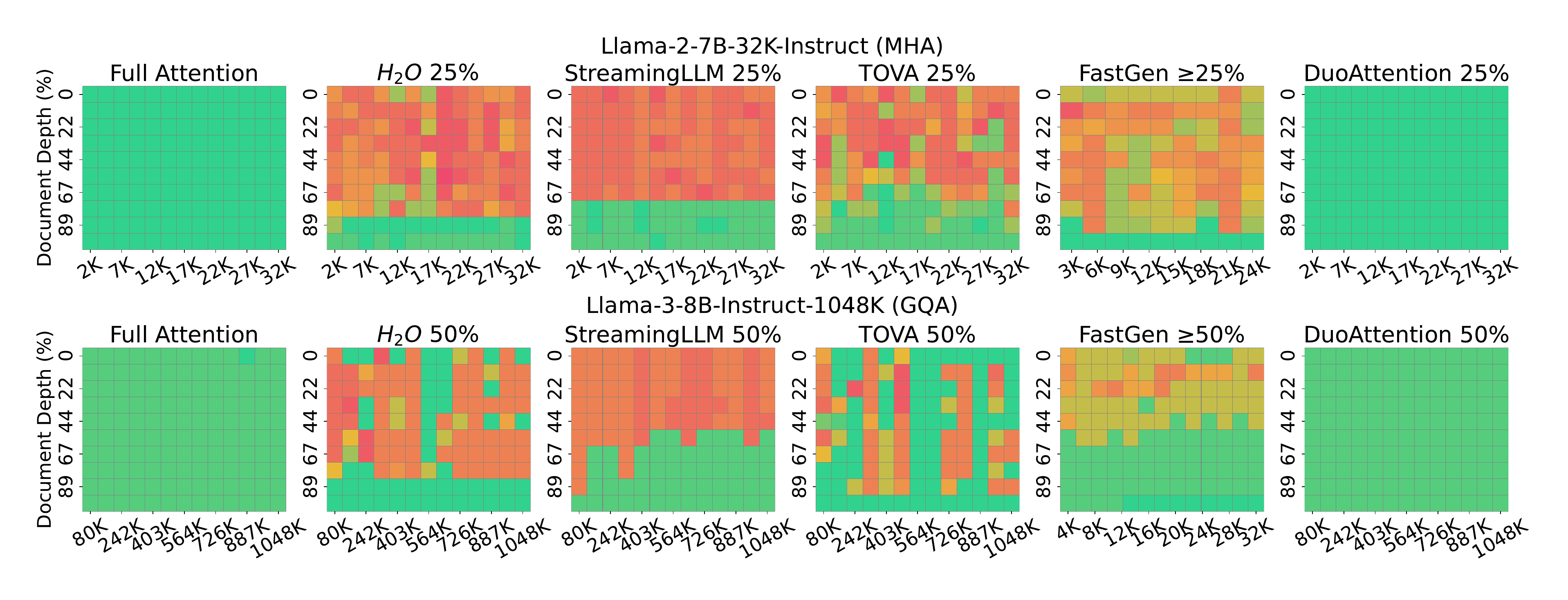}
    \caption{\small \method provides comparable accuracy as full attention on the Needle-in-a-Haystack benchmark using 25\% full attention ratio on the MHA model and 50\% full attention ratio on the GQA model.}
    \label{fig:nih_result}
\end{figure}

\begin{figure}[t]
    \centering
    \includegraphics[width=\textwidth]{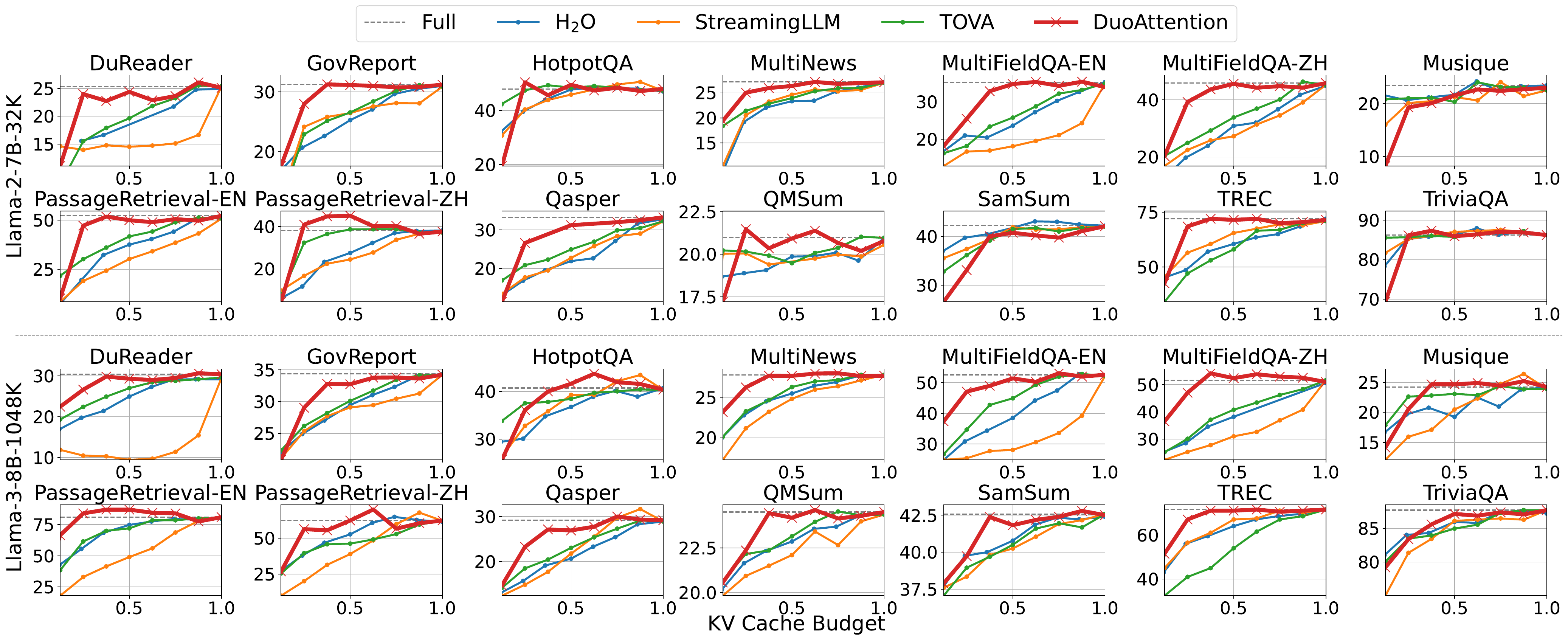}
    \caption{\small \method provides better KV budget and accuracy trade-off on LongBench benchmarks.}
    \label{fig:longbench}
\end{figure}

\section{Experiments}
\subsection{Setups}
\paragraph{Models, Datasets, and Baselines} We evaluate \method on both long-context and short-context benchmarks to demonstrate that our method preserves model performance on tasks requiring both long and short contexts while significantly improving efficiency. For long-context evaluations, we use the Needle-in-a-Haystack (NIAH) benchmark~\citep{LLMTest_NeedleInAHaystack} and LongBench~\citep{bai2023longbench}. For short-context evaluations, we assess performance on MMLU~\citep{hendryckstest2021}, MBPP~\citep{austin2021program}, and MT-Bench~\citep{zheng2023judging}. We employ state-of-the-art open-source models, including Llama-2-7B-chat~\citep{touvron2023llama2} (and its long-context variant Llama-2-7B-32K-Instruct~\citep{togetherlong}), Llama-3-[8,70]B-Instruct (and its long-context variant Llama-3-8B-Instruct-Gradient-1048k~\footnote{\url{https://huggingface.co/gradientai/Llama-3-8B-Instruct-Gradient-1048k}}), and Mistral-7B-v0.2-Instruct~\citep{jiang2023mistral}. We compare our method against KV cache compression algorithms, including H2O~\citep{zhang2023h2o}, TOVA~\citep{oren2024transformers}, FastGen~\citep{ge2024model}, and StreamingLLM~\citep{xiao2023streamingllm}.

\paragraph{Implementation Details} We implement \method in PyTorch~\citep{pytorch} using RoPE~\citep{su2021roformer} and RMSNorm kernels from FlashInfer~\citep{flashinfer}. For retrieval head identification, we use a batch size of 1, inserting ten 32-word passkeys into the BookSum~\citep{kryscinski2021booksum} dataset. The identification process uses 128 sink tokens and 256 recent tokens. Training samples are drawn from 50 intervals ranging from 1,000 tokens to the model-specific maximum length. Passkeys are randomly inserted at 1000 points within the context. Further details are included in Appendix Section~\ref{sec:experimental_details}.
We optimize gate values using the AdamW~\citep{DBLP:journals/corr/KingmaB14} optimizer, starting with a learning rate of 0.02, warming up from 0.002 in the first 400 steps, and reducing back to 0.002 in the final 400 steps. All experiments run for 2,000 steps on NVIDIA A100 GPUs.

\subsection{Long-Context Benchmarks}
We evaluate \method using the Needle-in-a-Haystack (NIAH) benchmark and LongBench~\citep{bai2023longbench}. We use two long-context models: Llama-2-7B-32K-Instruct and Llama-3-8B-Instruct-Gradient-1048k. We configure \method with a 25\% retrieval head ratio for Llama-2-7B-32K-Instruct and a 50\% ratio for Llama-3-8B-Instruct-Gradient-1048k. We compare \method with H2O, TOVA, and StreamingLLM using the same KV cache budget. We use 64 sink, 256 recent tokens, and 32,000 pre-filling chunk size for \method. Since the original designs of H2O and TOVA do not support long contexts, we modify their algorithms by replacing the pre-filling stage with FlashAttention and simulating decoding for the last 50 tokens of the input, following~\citet{tang2024quest}. FastGen's algorithm does not allow for the specification of the KV compression ratio, as it fluctuates with inputs. Therefore, we adjust the attention recovery ratio to ensure the KV cache budget is, on average, above 25\% or 50\% in the experiments shown in Figure~\ref{fig:nih_result}. Additionally, FastGen’s quadratic memory cost during the attention profiling phase limits its ability to handle long-context samples. We measure FastGen's performance on NIAH for Llama-2-7B up to a 24K context and for Llama-3-8B up to a 32K context; beyond these sizes, it results in out-of-memory errors. Detailed baseline implementations and justifications are provided in Appendix Section~\ref{sec:h2o_tova_implementation} and Section~\ref{sec:fastgen_implementation}.

\textbf{Needle-in-a-Haystack (NIAH)} is a challenging pressure test designed to assess the ability of models to accurate identify and retrieve relevant information from lengthy context. As shown in Figure~\ref{fig:nih_result}, all baseline methods fail to retrieve correct answers from the various depths of the long sequence, as they discard the KV cache containing the necessary information during generation. In contrast, \method retains all KV caches in the retrieval heads while discarding only those in the streaming heads, preserving the model's retrieval capability. As a result, \method demonstrates strong performance across all sequence depths, handling lengths up to 1048K tokens effectively.

\textbf{LongBench}~\citep{bai2023longbench} is a comprehensive suite of long-context datasets encompassing multiple tasks and natural texts, designed to assess long-context understanding capabilities more thoroughly. Figure~\ref{fig:longbench} shows the performance on 14 LongBench tasks, comparing different methods based on their KV cache budgets. \method shows a superior trade-off between KV budget and accuracy on most tasks, underscoring its generalizability. Notably, \method achieves performance comparable to full attention on most tasks, using a 25\% KV cache budget for MHA and a 50\% KV cache budget for GQA, consistent with the results observed in the needle-in-a-haystack benchmark. We compare \method with FastGen in Table~\ref{tab:longbench_fastgen_llama3} and~\ref{tab:longbench_fastgen_llama2} in the Appendix. 
Table~\ref{tab:longbench_llama3} and~\ref{tab:longbench_llama2} in the Appendix provides full results for all 21 LongBench tasks using the 25\% and 50\% KV cache budget for the two models, showing that \method consistently outperforms baselines across most tasks and achieves the highest average scores.

\begin{figure}[t]
    \begin{minipage}[t]{0.53\textwidth}\vspace{0pt}
        \centering
        \small
        \includegraphics[width=1.0\textwidth]{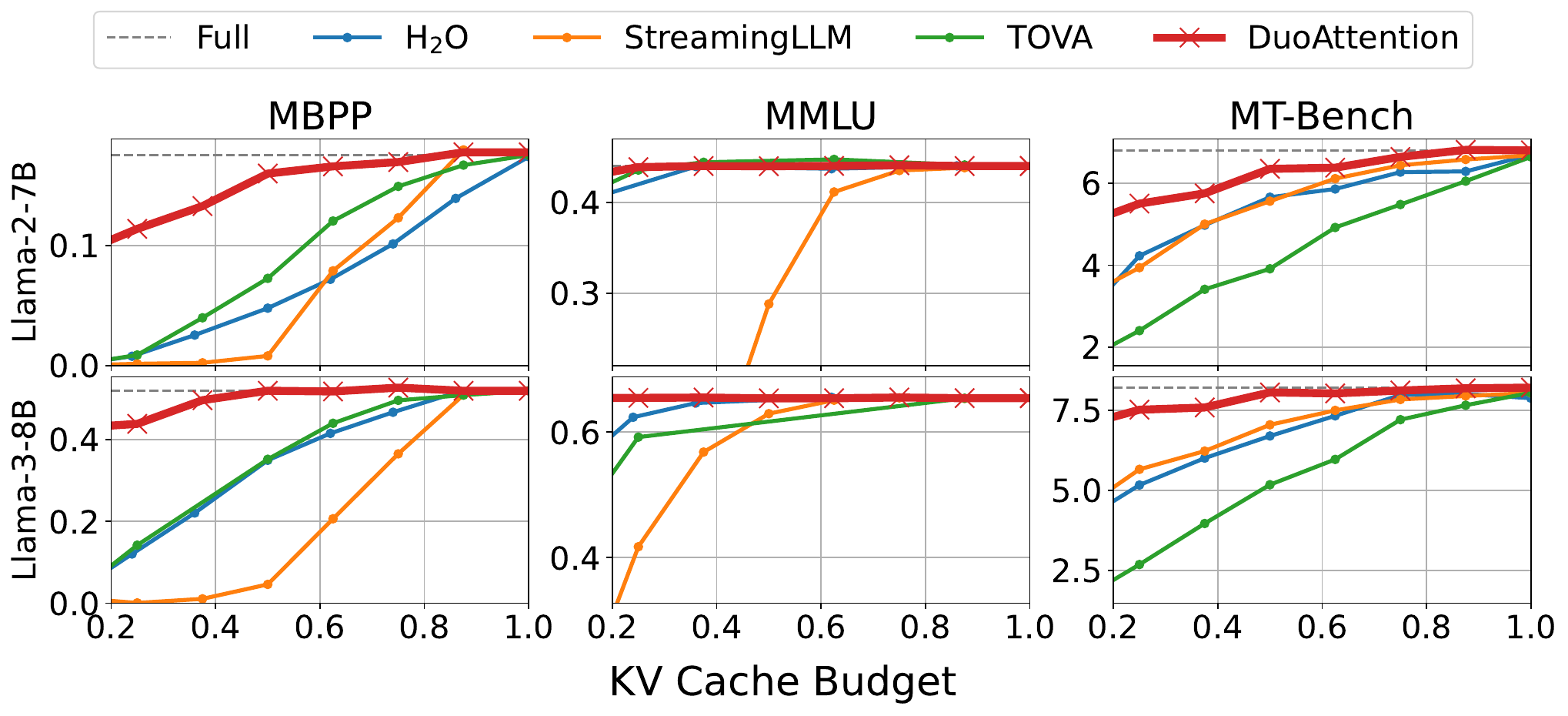}
        \captionof{figure}{\small Results on short benchmarks.}
        
        \label{fig:MBPP}
    \end{minipage}
    \hfill
    \begin{minipage}[t]{0.47\textwidth}\vspace{0pt}
        \centering
        \small
        \setlength{\tabcolsep}{1.5pt}
        \captionof{table}{\small Llama-3-70B results on short benchmarks.}
        \label{tab:llama-3-70B-short}
        \begin{tabular}{lcccc}
            \toprule
                                & Budget & MMLU             & MBPP             & MT-B          \\
            \midrule
            Full                & 100\%  & 79.38\%          & 47.85\%          & 8.93          \\
            \midrule
            H2O                 & 50\%   & 79.26\%          & 32.12\%          & 7.16          \\
            TOVA                & 50\%   & 79.15\%          & 36.09\%          & 7.96          \\
            SLLM                & 50\%   & 77.46\%          & 5.57\%           & 5.41          \\
            \textbf{DuoAttn} & 50\%   & \textbf{79.35\%} & \textbf{47.09\%} & \textbf{9.14} \\
            \bottomrule
        \end{tabular}
    \end{minipage}
\end{figure}

\subsection{Short-Context Benchmarks.}
To ensure that \method does not compromise the model's performance on short-context tasks, we evaluate it alongside all baselines on three short-context benchmarks: MMLU, MBPP, and MT-Bench. These benchmarks assess the model's knowledge, coding abilities, and helpfulness.
We use one-shot prompting for MMLU and zero-shot prompting for MBPP and MT-Bench. For \method, we configure 32 sink tokens and 128 recent tokens on MMLU, and 16 sink tokens and 64 recent tokens on MBPP and MT-Bench.
As shown in Figure~\ref{fig:MBPP} and Table~\ref{tab:llama-3-70B-short}, \method consistently outperforms all baselines under the same KV cache budget across various models, including Llama-2-7B, Llama-3-8B, and Llama-3-70B-Instruct. With a 50\% KV cache budget, \method achieves near-lossless performance on most benchmarks, demonstrating that it preserves the model's original capabilities.

\begin{figure}[t]
    \centering
    \includegraphics[width=\textwidth]{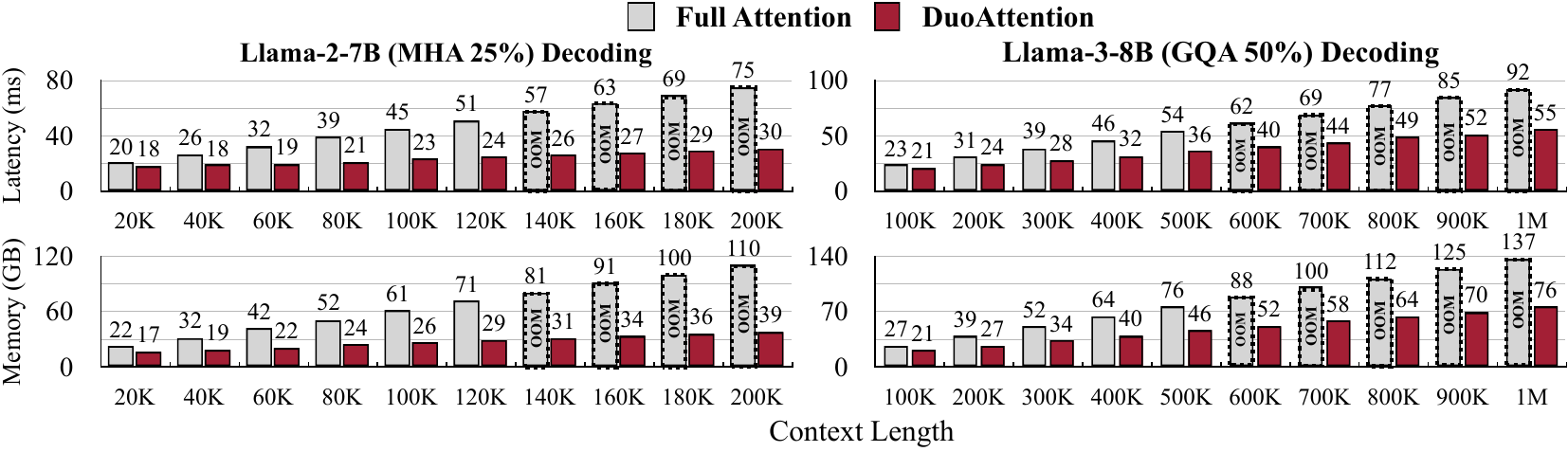}
    \caption{\small \textbf{Per-token decoding latency and memory} usage of \method compared to full attention across varying \textbf{context sizes}. \method uses a 25\% retrieval head ratio for Llama-2-7B (MHA) and 50\% for Llama-3-8B (GQA). \method achieves up to 2.45$\times$ memory reduction for MHA and 1.65$\times$ for GQA models, along with up to 2.13$\times$ latency reduction for MHA and 1.5$\times$ for GQA models. These reductions approach the inverse of the retrieval head ratios as context length increases. Out-of-memory (OOM) results are linearly extrapolated from measured data.}
    
    \label{fig:efficiency_decoding}
\end{figure}
\begin{figure}[t]
    \centering
    \includegraphics[width=\textwidth]{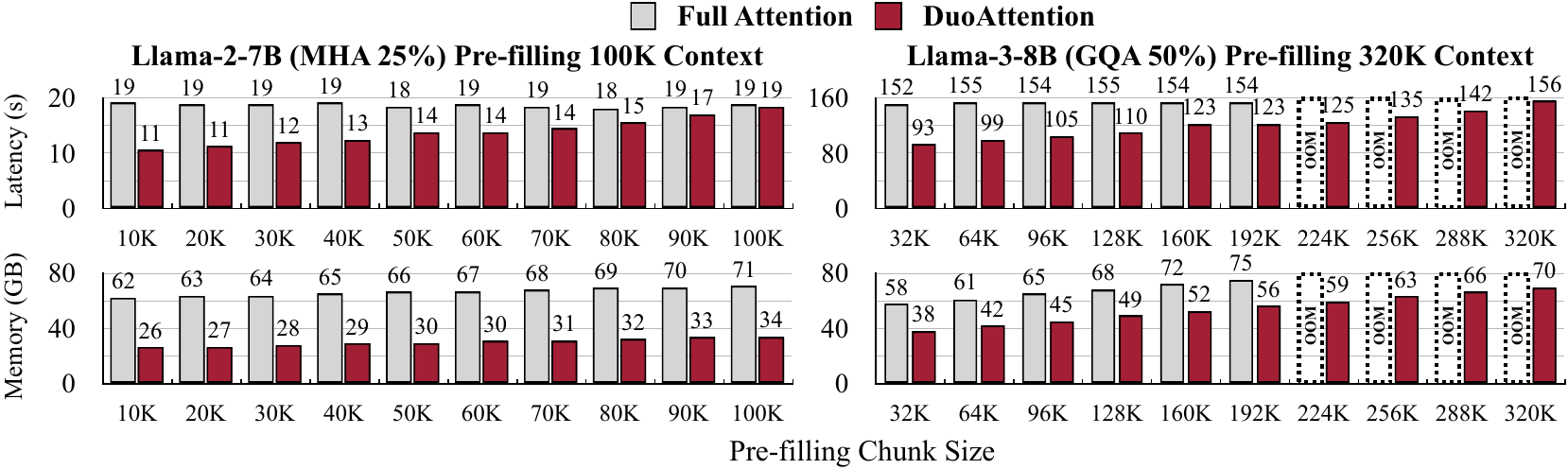}
    \caption{\small \textbf{Pre-filling latency and memory} usage of \method compared to full attention across varying \textbf{pre-filling chunk sizes}. \method uses a 25\% retrieval head ratio for Llama-2-7B (MHA), pre-filling a context of 100K tokens, and a 50\% ratio for Llama-3-8B (GQA), pre-filling a context of 320K tokens. As the pre-filling chunk size decreases, \method achieves up to 1.73$\times$ latency reduction for MHA and 1.63$\times$ for GQA models, with memory reductions up to 2.38$\times$ for MHA and 1.53$\times$ for GQA models.}
    \label{fig:efficiency_prefilling}
\end{figure}

\begin{figure*}[t]
    \centering
    \begin{minipage}{0.60\textwidth}
        \centering
        \includegraphics[width=\linewidth]{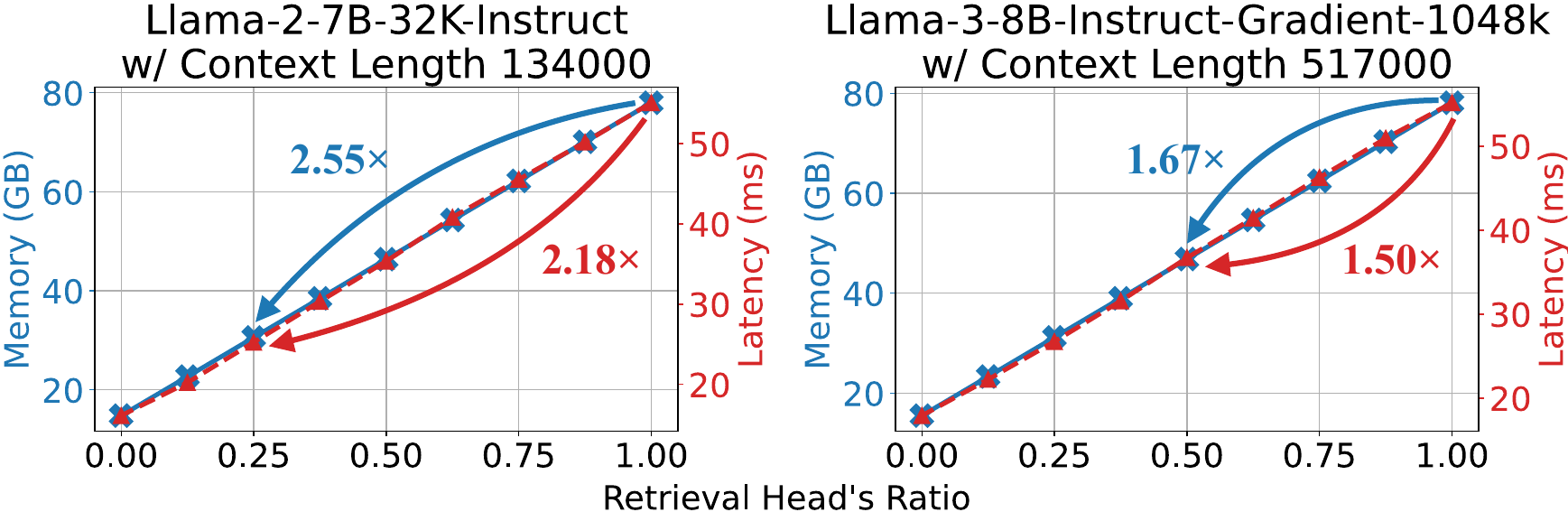}
        \vspace{-1em}
        \caption{\small \method's decoding memory and latency \vs KV budget with a fixed context length. Memory and latency are reduced linearly when the ratio of retrieval heads is reduced. \method achieves up to 2.55$\times$ memory reduction for MHA and 1.67$\times$ for GQA models, along with up to 2.18$\times$ latency reduction for MHA and 1.50$\times$ for GQA models.}
        \label{fig:efficiency_curve}
    \end{minipage}\hfill
    \begin{minipage}{0.38\textwidth}
        \centering
        \includegraphics[width=0.98\textwidth]{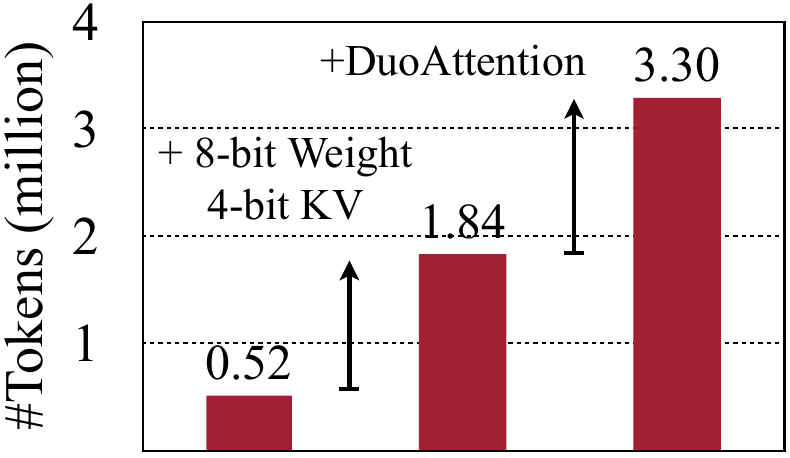}
        \caption{\small Combined with 8-bit weight and 4-bit KV cache quantization, \method can accommodate 3.30 million tokens on a single A100-80G GPU for the Llama-3-8B model.}
        \label{fig:kv_capacity}
    \end{minipage}
\end{figure*}

\subsection{Efficiency Results}
We evaluate \method's decoding latency and memory usage on Llama-2-7B and Llama-3-8B models on a single NVIDIA A100 GPU. We pre-allocate the KV cache for the entire benchmark sequence to prevent the extra overheads of dynamic memory allocations. The default number format for weights and activations is BFloat16. By employing a retrieval head ratio of 25\% for Llama-2-7B and 50\% for Llama-3-8B, \method maintains accuracy while significantly improving efficiency.
\myparagraph{Decoding Efficiency}
As shown in Figure~\ref{fig:efficiency_decoding}, \method's decoding speed scales linearly, though with a flatter slope compared to full attention, reflecting the chosen retrieval head ratio. This efficient scaling leads to significant reductions in memory usage and notable improvements in decoding speed. These improvements approach the inverse of the retrieval head ratios as context length increases.
Figure~\ref{fig:efficiency_curve} shows \method's speedup and memory savings across various KV budget settings for a fixed context size. Both decoding latency and memory usage decrease linearly as the ratio of retrieval heads is reduced in the deployment configuration. Under the settings in Figure~\ref{fig:efficiency_curve}, \method achieves maximum improvements on an A100 GPU: 2.55$\times$ memory reduction for MHA and 1.67$\times$ for GQA models, and 2.18$\times$ latency reduction for MHA and 1.50$\times$ for GQA models.

\myparagraph{Pre-filling Efficiency} \method also accelerates long-context pre-filling for LLMs, as discussed in Section~\ref{sec:prefilling}. 
Figure~\ref{fig:efficiency_prefilling} shows that \method significantly reduces both pre-filling latency and memory usage, with these savings increasing as the pre-filling chunk size decreases. This is because the time and memory complexity for the streaming heads are reduced with smaller chunk sizes. \method achieves up to 1.73$\times$ latency reduction for MHA and 1.63$\times$ for GQA models, with memory reductions of up to 2.38$\times$ for MHA and 1.53$\times$ for GQA models.

\myparagraph{Combiniation with Quantization}
To fit more tokens into limited memory, we can integrate weight and KV cache quantization with \method to maximize KV cache capacity. Previous studies have shown that weight quantization~\citep{xiao2023smoothquant,lin2024awq} and 4-bit KV cache quantization~\citep{lin2024qserve,liu2024kivi,hooper2024kvquant} do not compromise model performance. We combine \method with the QServe~\citep{lin2024qserve} quantization method and kernels to enable 8-bit weight and 4-bit KV cache LLM inference.
Measured results are shown in Figure~\ref{fig:kv_capacity}. Combining quantization techniques with \method allows us to accommodate up to 3.30 million tokens on a single A100-80G GPU using the Llama-3-8B model, resulting in a 6.4$\times$ increase in capacity compared to the naive full attention BF16 deployment.

\subsection{Ablation Studies}
\label{sec:ablation}
\begin{figure}[t]
    \centering
    \includegraphics[width=1.0\textwidth]{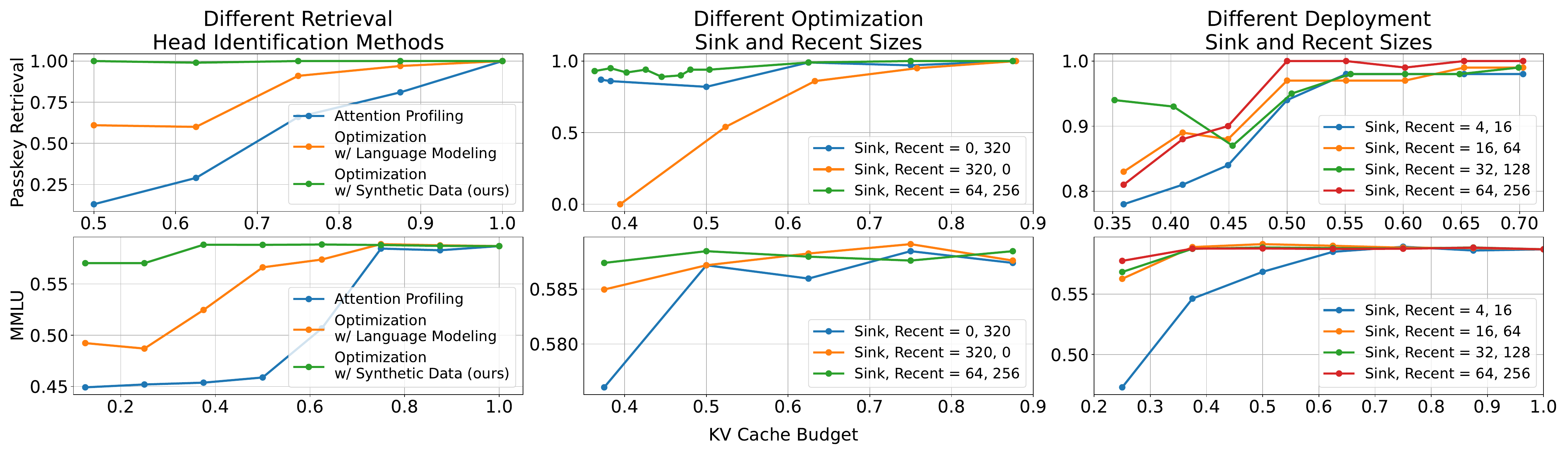}
    \caption{\small Ablation studies: (1) Comparison of retrieval head identification methods, showing the superiority of our optimization-based approach with synthetic data over attention profiling and language modeling. (2) Analysis of start and recent token sizes shows that combining sink and recent attention optimally identifies retrieval heads. (3) Deployment performance indicates 16 attention sinks and 64 recent tokens are optimal, with minimal gains beyond these values.}
    \label{fig:ablations}
\end{figure}
We conduct ablation studies using the Mistral-7B-Instruct-v0.2 on passkey retrieval and MMLU datasets. For the passkey retrieval task, we embed an 8-word passkey within a 30K-word text and perform a linear sweep across 100 insertion depths, reporting exact match accuracies.
\paragraph{Optimization-based vs. Attention Profiling-based Retrieval Head Identification}
We assess our optimization-based method against attention profiling, as used in FastGen~\citep{ge2024model} and RazorAttention~\citep{tang2024razorattentionefficientkvcache}, utilizing the same synthetic passkey dataset for both. Results in Figure~\ref{fig:ablations} (1) show our method significantly outperforms attention profiling, which struggles to identify retrieval heads, affecting model optimization accurately.

\paragraph{Optimizing with Synthetic Data vs. Language Modeling}
As illustrated in Figure~\ref{fig:ablations} (1), our approach of using synthetic data to identify retrieval heads produces significantly better results than traditional language modeling, which computes loss on all tokens in natural data.

\paragraph{Necessity of Sink+Recent Attention in Optimization}
Figure~\ref{fig:ablations} (2) highlights the importance of combining sink and recent attention during the optimization phase. Exclusive reliance on either starting or recent token attention is inadequate for effective retrieval head identification.

\paragraph{Deployment Phase Configuration}
We analyze the deployment configuration for attention sinks and recent tokens within streaming heads. Our findings indicate that performance plateaus at 16 sink tokens and 64 recent tokens (Figure~\ref{fig:ablations} (3)). Further increases yield marginal improvements.

\section{Related Work}
Various approaches have been developed to scale up LLMs and improve their efficiency in handling long contexts. These methods can be grouped into four main categories: optimizing model architectures, using approximate attention mechanisms, applying KV cache quantization, and system-level optimizations.

\myparagraph{Model Architecture} Multi-Query Attention (MQA)\citep{shazeer2019fast} and Grouped-Query Attention (GQA)\citep{ainslie2023gqa} reduce the size of the Key-Value (KV) cache by sharing KV heads across query heads. However, these methods require pre-training with specific architectures and do not reduce computational costs. Linear attention Transformers~\citep{gu2023mamba} reduce memory usage but tend to underperform on tasks requiring long-context processing.

\myparagraph{Approximate Attention} Methods like Sparse Transformer~\citep{child2019generating} and LongFormer~\citep{beltagy2020longformer} use local or block attention patterns to reduce computational complexity. BigBird~\citep{zaheer2020big} achieves linear complexity by combining local and global attention, but many of these methods require custom GPU kernels or retraining, limiting their practicality. H2O~\citep{zhang2023h2o} and TOVA~\citep{oren2024transformers} simplify attention by discarding tokens based on query patterns. StreamingLLM~\citep{xiao2023streamingllm} identifies "attention sinks" and proposes always retaining initial and recent tokens to maintain constant decoding latency and memory usage, allowing the model to process significantly more input tokens than the pre-training sequence length. FastGen~\citep{ge2024model} profiles attention heads to discard tokens during decoding. However, our experiments show that these methods degrade the long-context abilities of LLMs. Also, these methods cannot reduce the pre-filling cost of long-context LLMs.

\myparagraph{KV Cache Quantization} Techniques such as 8-bit and 4-bit quantization~\citep{liu2024kivi,hooper2024kvquant,lin2024qserve} reduce the size of KV caches, but they do not address the computational overhead of attention kernels. These methods are complementary to \method and can be used together to further reduce memory usage.

\myparagraph{System Optimizations} vLLM~\citep{kwon2023efficient} and FlashAttention~\citep{dao2022flashattention,dao2023flashattention2} improve attention computation efficiency by optimizing batch processing and utilizing GPU memory hierarchies. FlashDecoding~\citep{hong2024flashdecoding} and RingAttention~\citep{liu2023ring} introduce further improvements in decoding speed and sequence-level parallelism. While these methods enhance computational performance, they do not address KV cache size reduction, making them complementary to \method for additional speed and memory optimization.

\myparagraph{Recent Works} Several recent works share similar ideas with \method. \citet{wu2024retrieval} introduces the concept of retrieval heads to explain LLMs' long-context capabilities. However, their approach does not compress the KV cache for non-retrieval heads, focusing solely on accuracy. MInference~\citep{jiang2024minference} accelerates pre-filling for long-context LLMs by using sparse attention patterns but does not optimize KV cache storage or latency during decoding. RazorAttention~\citep{tang2024razorattentionefficientkvcache} also divides attention heads into retrieval and non-retrieval categories but relies on attention profiling, which, as our experiments show, is less accurate than our optimization-based approach. Also, RazorAttention doesn't optimize pre-filling. \method offers more effective KV cache management and higher compression rates, leading to better performance for both pre-filling and decoding in long-context applications.
\section{Conclusion}
We introduce \method, a framework that optimizes memory and computational resources in LLMs by distinguishing between \emph{Retrieval Heads} and \emph{Streaming Heads}. By applying a full KV cache only to retrieval heads, \method significantly reduces memory usage and latency for both decoding and pre-filling in long-context applications. It achieves memory reductions of up to 2.55$\times$ for MHA and 1.67$\times$ for GQA models, with decoding speed improvements of up to 2.18$\times$ for MHA and 1.50$\times$ for GQA, and pre-filling accelerations of up to 1.73$\times$ and 1.63$\times$, respectively, with minimal accuracy loss compared to full attention.
When combined with quantization, \method further boosts KV cache capacity, supporting up to 3.30 million contextual tokens on a single A100 GPU. \method paves the way for LLMs to handle contexts with millions of tokens.

\subsubsection*{Acknowledgments}
We thank MIT-IBM Watson AI Lab, MIT and Amazon Science Hub, MIT AI Hardware Program, National Science Foundation, Hyundai and Samsung for supporting this research. We thank NVIDIA for donating the DGX server.  

\bibliography{reference}
\bibliographystyle{iclr2025_conference}

\clearpage
\appendix
\section{Appendix}
\subsection{Experimental Details}
\label{sec:experimental_details}
\begin{figure}[h]
    \centering
    \includegraphics[width=0.3\linewidth]{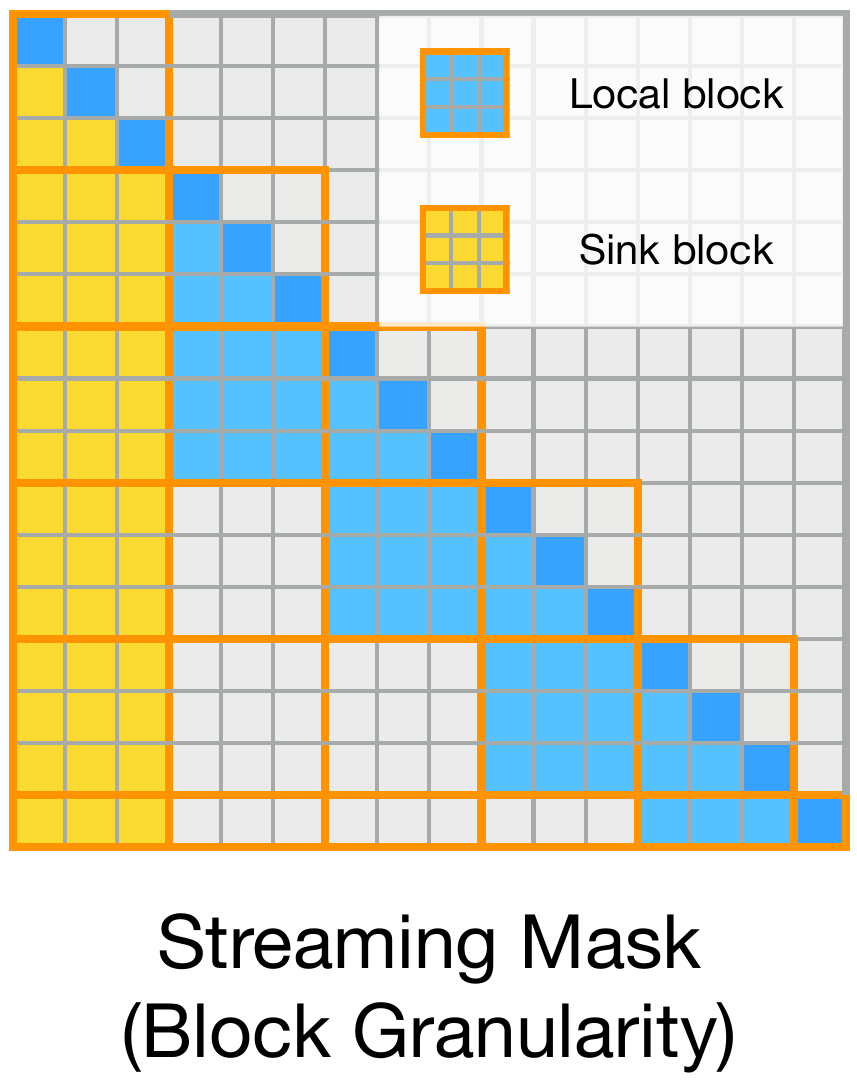}
    \caption{\small Block-sparse approximation of $\Lambda$-like attention.}
    \label{fig:block-streaming}
\end{figure}

We use FSDP2 in PyTorch for model training and DeepSpeed Ulysses~\citep{jacobs2023deepspeedulyssesoptimizationsenabling} sequence parallelism to support long sequences. During training, we use an efficient block-sparse approximation of $\Lambda$-like attention for streaming attention, as implemented in~\citet{guo2024blocksparse} and illustrated in Figure~\ref{fig:block-streaming}. Maximum sequence lengths vary across models, as detailed in Table~\ref{tab:hyperparams}.
\begin{table}[h]
    \centering
    \caption{\small Training Hyperparameters.}
    \label{tab:hyperparams}
    \begin{tabular}{lc}
        \toprule
        Models & Max. Seq. Lengths\\
        \midrule
        Llama-2-7B-chat & 4096\\
        Llama-2-7B-32K-Instruct & 32000\\
        Llama-3-8B-Instruct & 8192\\
        Llama-3-8B-Instruct-1048K & 32000\\
        Llama-3-70B-Instruct & 8192\\
        Mistral-7B-Instruct-v0.2 & 32000\\
        \bottomrule
    \end{tabular}
\end{table}

\subsection{Full LongBench Results}

\begin{table}[h]
    \centering
    \caption{Full LongBench results with Llama-3-8B-Instruct-1048K. \method achieves the best performance with a 50\% KV cache budget on most datasets.}
    \label{tab:longbench_llama3}
    \begin{tabular}{lccccc}
        \toprule
        Dataset             & {Full} & {H2O (50\%)}   & {SLLM (50\%)} & {TOVA (50\%)}  & \textbf{Duo (50\%)} \\
        \midrule
        \textbf{Average}             & 40.08 & 35.76 & 32.26 & 35.55 & \textbf{40.21} \\
        \midrule
        2WikiMQA            & 28.78 & 27.99 & \textbf{29.22} & 26.93 & 29.08 \\
        DuReader (zh)       & 30.41 & 24.94 & 9.41  & 27.00 & \textbf{29.31} \\
        GovReport           & 34.23 & 29.44 & 29.08 & 30.10 & \textbf{32.72} \\
        HotpotQA            & 40.37 & 36.77 & 39.27 & 38.45 & \textbf{41.63} \\
        LCC                 & 38.19 & 43.09 & 41.94 & 42.31 & \textbf{44.16} \\
        LSHT (zh)           & 38.00 & 25.00 & 25.50 & 24.50 & \textbf{30.00} \\
        MultiNews           & 27.73 & 25.52 & 24.85 & 26.32 & \textbf{27.72} \\
        MultiFieldQA-en     & 52.62 & 38.53 & 28.11 & 44.94 & \textbf{51.44} \\
        MultiFieldQA-zh     & 50.58 & 38.25 & 31.07 & 40.82 & \textbf{52.40} \\
        Musique             & 24.22 & 19.24 & 20.47 & 23.07 & \textbf{24.65} \\
        NarrativeQA         & 26.56 & 25.13 & 22.06 & \textbf{25.64} & 24.54 \\
        Passage Count       & 1.00  & \textbf{2.05}  & 1.64  & 1.00  & 0.00  \\
        PassageRetrieval-en & 81.00 & 74.75 & 49.00 & 72.00 & \textbf{87.00} \\
        PassageRetrieval-zh & 62.15 & 52.57 & 38.90 & 46.13 & \textbf{62.15} \\
        Qasper              & 29.21 & 20.65 & 21.77 & 23.06 & \textbf{26.93} \\
        QMSum               & 24.52 & 22.87 & 22.11 & 23.16 & \textbf{24.20} \\
        RepoBench-P         & 38.94 & 39.98 & 37.60 & 40.14 & \textbf{46.12} \\
        SAMSum              & 42.51 & 40.78 & 40.25 & 40.50 & \textbf{41.83} \\
        TREC                & 71.50 & 64.00 & 67.00 & 54.00 & \textbf{71.00} \\
        TriviaQA            & 87.70 & 85.98 & 86.11 & 84.97 & \textbf{87.14} \\
        VCSUM (zh)          & 11.37 & \textbf{13.45} & 12.10 & 11.59 & 10.46 \\
        \bottomrule
    \end{tabular}
\end{table}

\begin{table}[h]
    \centering
    \caption{Full LongBench results with Llama-2-7B-Instruct-32K. \method achieves the best performance with a 25\% KV cache budget on most datasets.}
    \label{tab:longbench_llama2}
    \begin{tabular}{lccccc}
        \toprule
        Dataset             & {Full} & {H2O (25\%)}   & {SLLM (25\%)} & {TOVA (25\%)}  & \textbf{Duo (25\%)} \\
        \midrule
        \textbf{Average}             & 37.52 & 26.84 & 27.80 & 29.78 & \textbf{34.49} \\
        \midrule
        2WikiMQA            & 35.59 & 28.87 & 29.69 & 31.18 & \textbf{33.37} \\
        DuReader (zh)       & 25.10 & 15.56 & 13.96 & 15.51 & \textbf{23.99} \\
        GovReport           & 31.23 & 20.66 & 24.14 & 22.88 & \textbf{27.98} \\
        HotpotQA            & 47.98 & 39.60 & 40.39 & 47.45 & \textbf{50.44} \\
        LCC                 & 51.21 & 45.78 & 44.25 & 47.91 & \textbf{48.34} \\
        LSHT (zh)           & 34.50 & 16.50 & 17.50 & 18.50 & \textbf{25.50} \\
        MultiNews           & 27.11 & 19.21 & 20.54 & 21.41 & \textbf{25.03} \\
        MultiFieldQA-en     & 33.95 & 21.01 & 16.69 & 18.19 & \textbf{25.49} \\
        MultiFieldQA-zh     & 45.79 & 19.81 & 22.50 & 24.96 & \textbf{39.23} \\
        Musique             & 22.97 & 20.63 & 20.09 & \textbf{21.00} & 19.27 \\
        NarrativeQA         & 24.11 & 19.14 & 21.13 & \textbf{23.06} & 20.49 \\
        Passage Count       & 0.00  & 0.53  & \textbf{0.58}  & 0.00  & 0.33  \\
        PassageRetrieval-en & 50.92 & 19.50 & 19.08 & 30.17 & \textbf{47.25} \\
        PassageRetrieval-zh & 37.68 & 11.75 & 16.77 & 32.38 & \textbf{40.93} \\
        Qasper              & 33.23 & 16.84 & 17.68 & 20.85 & \textbf{26.59} \\
        QMSum               & 20.79 & 18.89 & 20.05 & 20.16 & \textbf{21.48} \\
        RepoBench-P         & 51.58 & 45.16 & 45.25 & \textbf{49.03} & 48.58 \\
        SAMSum              & 42.10 & \textbf{39.73} & 37.43 & 36.17 & 33.10 \\
        TREC                & 71.50 & 48.50 & 56.50 & 47.00 & \textbf{68.50} \\
        TriviaQA            & 86.21 & 85.16 & 85.24 & 85.65 & \textbf{86.15} \\
        VCSUM (zh)          & 14.45 & 10.71 & \textbf{14.36} & 11.85 & 12.35 \\
        \bottomrule
    \end{tabular}
\end{table}

\begin{table}[h]
    \centering
    \caption{Comparison of FastGen and \method on a subset of LongBench using the Llama-3-8B-Instruct-1048K model.}
    \label{tab:longbench_fastgen_llama3}
    \begin{tabular}{lcc}
        \toprule
                    & FastGen (\textgreater{}50\%) & \textbf{DuoAttention (50\%)} \\
                    \midrule
Average             & 32.82                                            & \textbf{40.01}                                   \\
\midrule
2WikiMQA            & 18.61                                            & \textbf{29.08}                                   \\
DuReader (zh)       & 20.22                                            & \textbf{29.31}                                   \\
HotpotQA            & 33.08                                            & \textbf{41.63}                                   \\
LCC                 & \textbf{46.50}                                            & 44.16                                   \\
MultiNews           & 18.18                                            & \textbf{27.72}                                   \\
MultiFieldQA-en     & 44.05                                            & \textbf{51.44}                                   \\
MultiFieldQA-zh     & 42.15                                            & \textbf{52.40}                                   \\
Musique             & 13.58                                            & \textbf{24.65}                                   \\
Passage Count       & 0.09                                             & 0.00                                    \\
PassageRetrieval-en & \textbf{93.12}                                            & 87.00                                   \\
PassageRetrieval-zh & 40.75                                            & \textbf{62.15}                                   \\
Qasper              & 26.51                                            & \textbf{26.93}                                   \\
QMSum               & 24.03                                            & \textbf{24.20}                                   \\
SAMSum              & 34.12                                            & \textbf{41.83}                                   \\
TriviaQA            & 69.92                                            & \textbf{87.14}                                   \\
VCSUM (zh)          & 0.23                                             & \textbf{10.46}                                   \\
        \bottomrule
    \end{tabular}
\end{table}

\begin{table}[h]
    \centering
    \caption{Comparison of FastGen and \method on a subset of LongBench using the Llama-2-7B-32K-Instruct model.}
    \label{tab:longbench_fastgen_llama2}
    \begin{tabular}{lcc}
        \toprule
                    & {FastGen (\textgreater{}25\%)} & \textbf{DuoAttention (25\%)} \\
                    \midrule
Average             & {19.01}                        & {\textbf{32.81}}               \\
\midrule
2WikiMQA            & {28.05}                        & {\textbf{33.37}}               \\
MultiNews           & {12.60}                        & {\textbf{25.03}}               \\
MultiFieldQA-en     & {\textbf{28.58}}                        & {25.49}               \\
MultiFieldQA-zh     & {22.44}                        & {\textbf{39.23}}               \\
PassageRetrieval-zh & {3.38}                         & {\textbf{40.93}}               \\
        \bottomrule
    \end{tabular}
\end{table}

\subsection{Implementation of H2O and TOVA on Long-Context Benchmarks}
\label{sec:h2o_tova_implementation}
The original designs of the H2O and TOVA algorithms are not compatible with FlashAttention during pre-filling, as they rely on attention scores to perform token eviction. Since attention scores in FlashAttention are never materialized, these algorithms cannot be used in pre-filling, which is one of their main flaws. Therefore, it's not possible to evaluate these algorithms in long-context settings like needle-in-the-haystack and LongBench, as they cause OOM during context pre-filling. To compare with these strategies, we modified the algorithms: during pre-filling, we used FlashAttention for exact calculations. During the decoding stage, we perform token eviction based on the generated tokens' attention scores to contextual tokens. This modification improves performance compared to the original design since pre-filling is exact and token eviction occurs only during decoding. In extreme scenarios, if there is only one generated token in the answer (e.g. multiple-choice tasks), our implementation of H2O and TOVA will be exact with full attention, unlike their true accuracy. To approach their true performance, we simulate the last 50 tokens in long input benchmarks (needle-in-the-haystack and LongBench) as generated tokens to perform their token eviction policy long enough, as well as our algorithm. This experimental setting is also used by ~\citet{tang2024quest}. Experimental results show our method can pass this pressure test, while H2O and TOVA cannot. 

\subsection{NIAH results on Mistral models}
\begin{figure*}[h]
    \centering
    \begin{minipage}{0.48\textwidth}
        \centering
        \includegraphics[width=\linewidth]{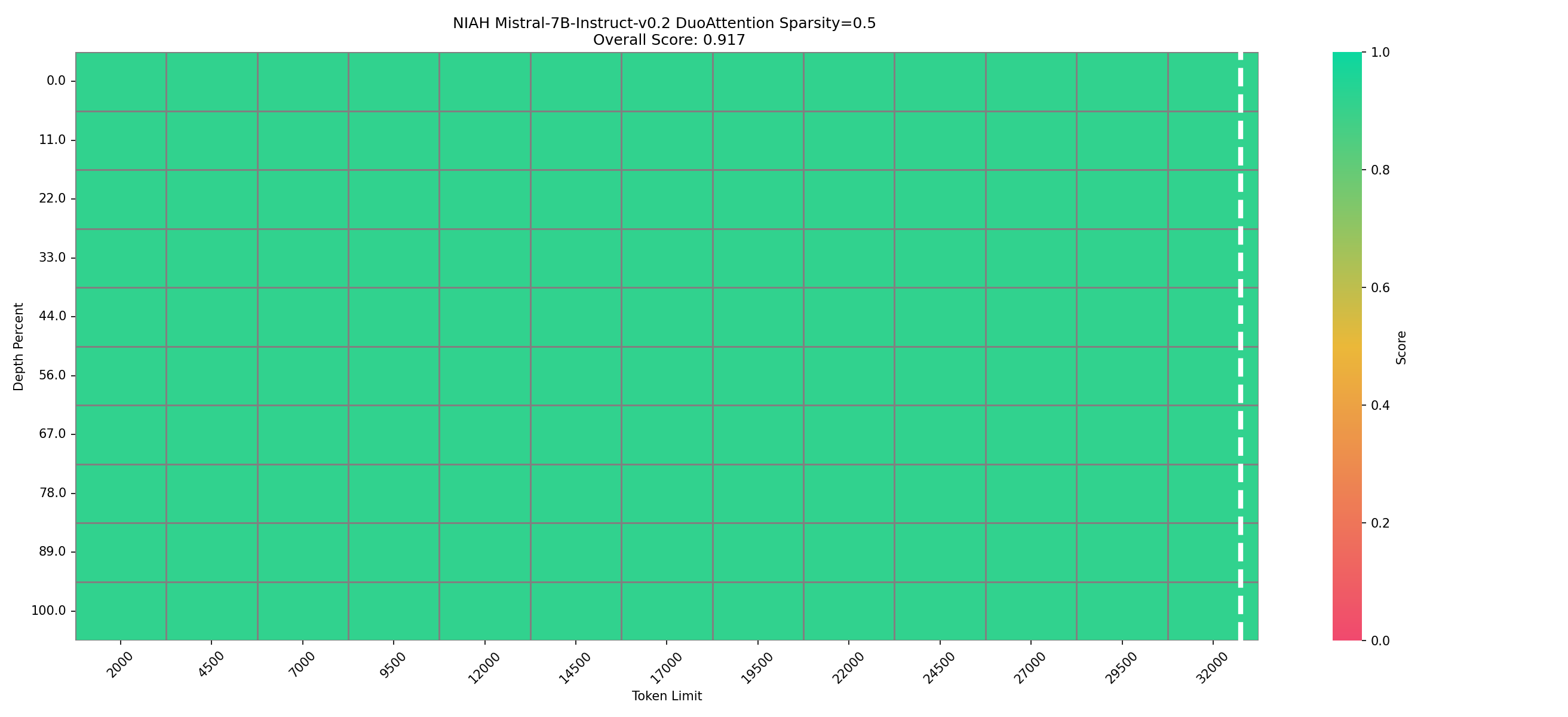}
        \caption{\small NIAH result on the Mistral-7B-Instruct-v0.2 model.}
        \label{fig:Mistral-0.2}
    \end{minipage}\hfill
    \begin{minipage}{0.48\textwidth}
        \centering
        \includegraphics[width=\linewidth]{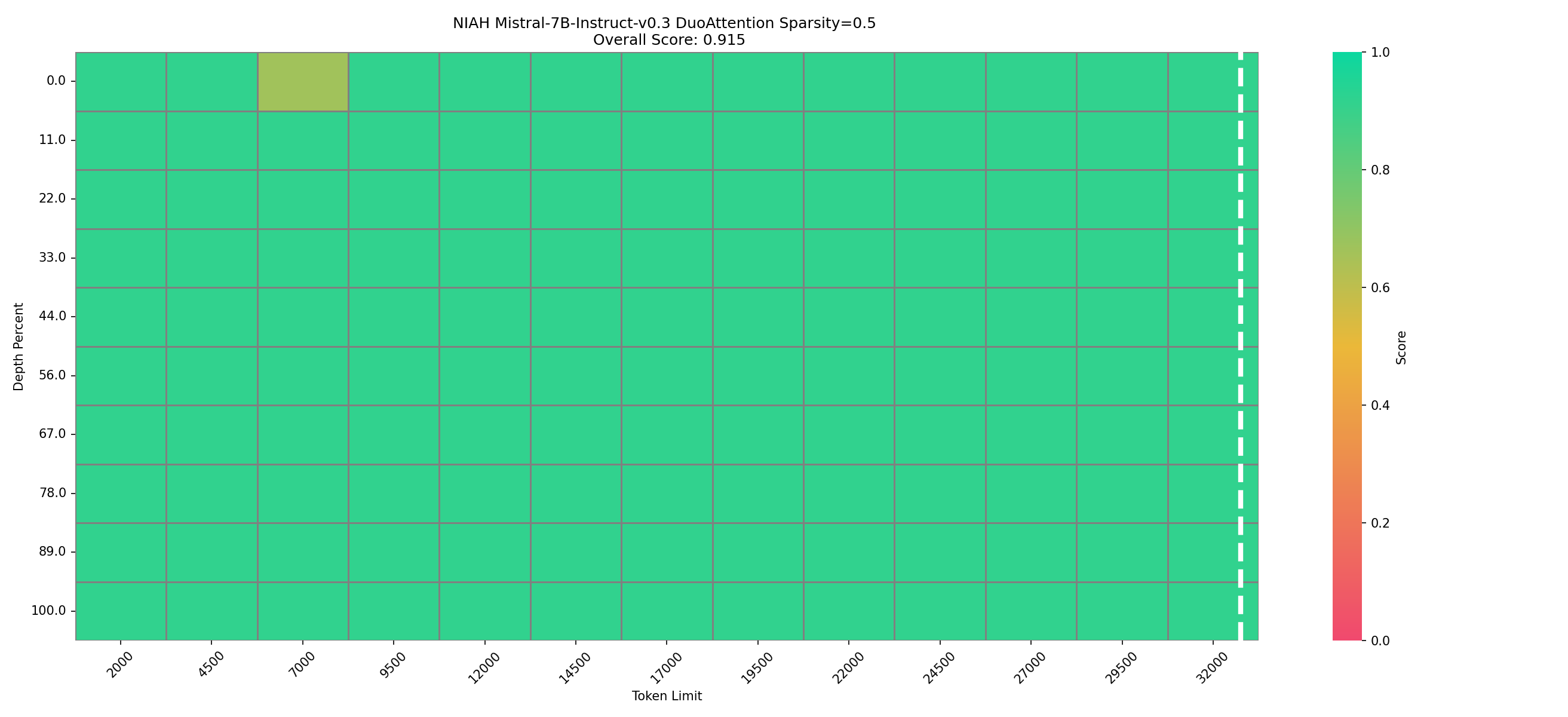}
        \caption{\small NIAH result on the Mistral-7B-Instruct-v0.3 model.}
        \label{fig:Mistral-0.3}
    \end{minipage}
\end{figure*}

\subsection{Implementation of FastGen on Long-Context Benchmarks}
\label{sec:fastgen_implementation}
Due to the lack of official implementation of the FastGen~(\cite{ge2024model}) algorithm, we reproduce it using a community codebase~(\cite{cold-compress-2024}), which is referenced by FastGen’s official repository. In the FastGen algorithm, the pruning ratio cannot be directly configurable; instead, the recovery ratio $T$ is used to control sparsity as outlined in the FastGen paper. To quantify sparsity, we calculated the average KV cache usage across all test cases as the overall measure of sparsity. For the Llama-2-7B model, we set the recovery ratio to $0.7$, ensuring the average KV cache budget was over 25\% of the full KV cache. Similarly, for the Llama-3-8B model, we set the recovery ratio to $0.87$, ensuring the average KV cache budget was more than 50\% of the full KV cache. Additionally, since FastGen uses the full attention map of the user-provided prompt to profile the types of different heads, it results in an $O(n^2)$ attention map complexity. Therefore, we are unable to test its performance in long contexts. For the long context benchmark, we used 8 A100-80G GPUs, achieving sequence lengths of up to 24k tokens for the Llama-2-7B model and up to 32k tokens for the Llama-3-8B model. In addition to the needle-in-the-haystack benchmark shown in Figure~\ref{fig:nih_result}, we also evaluated FastGen on LongBench for both models. However, due to the quadratic memory consumption of FastGen, we only report results for datasets that were feasible to run on 8x A100-80G GPUs using FastGen. As shown in Table \ref{tab:longbench_fastgen_llama3} and Table \ref{tab:longbench_fastgen_llama2}, DuoAttention can consistently outperform FastGen on LongBench datasets.

\end{document}